\pdfoutput=1

\documentclass[11pt]{article}

\usepackage[final]{acl}

\usepackage{times}
\usepackage{latexsym}

\usepackage[T1]{fontenc}

\usepackage[utf8]{inputenc}

\usepackage{microtype}

\usepackage{inconsolata}

\usepackage{graphicx}
\usepackage{multirow}

\usepackage{amsmath,amssymb} 
\usepackage{booktabs}      
\usepackage{url}           
\usepackage{xspace}        

\newcommand{\name}{Drift-Adapter\xspace}
\DeclareMathOperator*{\argmin}{arg\,min} 

\title{\name: A Practical Approach to Near Zero-Downtime Embedding Model Upgrades in Vector Databases}

\author{Harshil Vejendla \\
  Rutgers University - New Brunswick \\
  \texttt{harshil.vejendla@rutgers.edu}}

\begin{document}
\maketitle
\begin{abstract}
Upgrading embedding models in production vector databases typically necessitates re-encoding the entire corpus and rebuilding the Approximate Nearest Neighbor (ANN) index, leading to significant operational disruption and computational cost. This paper presents \name, a lightweight, learnable transformation layer designed to bridge embedding spaces between model versions. By mapping new queries into the legacy embedding space, \name enables the continued use of the existing ANN index, effectively deferring full re-computation. We systematically evaluate three adapter parameterizations: Orthogonal Procrustes, Low-Rank Affine, and a compact Residual MLP, trained on a small sample of paired old/new embeddings. Experiments on MTEB text corpora and a CLIP image model upgrade (1M items) show that \name recovers 95–99\% of the retrieval recall (Recall@10, MRR) of a full re-embedding, adding less than $10\,\mu\text{s}$ query latency. Compared to operational strategies like full re-indexing or dual-index serving, \name dramatically reduces recompute costs (by over $100\times$) and facilitates upgrades with near-zero operational interruption. We analyze robustness to varied model drift, training data size, scalability to billion-item systems, and the impact of design choices like diagonal scaling, demonstrating \name's viability as a pragmatic solution for agile model deployment.
\end{abstract}

\section{Introduction}\label{sec:intro}
Vector databases are foundational to modern retrieval, recommendation, and semantic search systems \citep{goodfellow2016deep}. These systems rely on embeddings generated by deep learning models, which are continuously improved. However, deploying an updated embedding model in a production environment presents a significant operational challenge: the entire corpus of stored items, potentially billions of vectors, must be re-encoded with the new model, and the corresponding ANN index rebuilt \citep{xu2023spfreshii}. This process is computationally intensive, time-consuming, and often leads to service downtime or periods of degraded performance. While this full re-computation might be desirable in the long term for optimal performance with the new model, the immediate operational burden is substantial.

This work addresses the practical challenge of minimizing this disruption. We investigate if a compact, efficient mapping can bridge successive embedding spaces, allowing services to leverage new models quickly while deferring the cost of a full corpus overhaul.
Building on embedding space alignment principles from cross-lingual and cross-domain research \citep{schonemann1966ags, xing2024crosslingualwe, gao2023vec2vecac}, we introduce and systematically evaluate \name, a lightweight adaptation layer. Trained on a small set of paired old/new embeddings, \name transforms queries encoded by the new model into the legacy space of the existing database, enabling direct querying of the unchanged ANN index (e.g., FAISS \citep{johnson2021billion}). This facilitates upgrades with near-zero operational interruption.

Our contributions are:
\begin{itemize}
    \item We frame the operational problem of model upgrades in vector databases as a specific embedding alignment task and perform a systematic benchmark of lightweight methods in this context, detailing the practical trade-offs (recall, latency, cost) between Orthogonal Procrustes, Low-Rank Affine, and Residual MLP adapters.
    \item We conduct extensive experiments on text (MTEB \citep{muennighoff2022mtebmt}) and image (LAION \citep{schuhmann2021laion400mod}) corpora, demonstrating 95-99\% recall recovery with minimal overhead.
    \item We benchmark \name against common operational upgrade strategies (full re-index, dual index), quantifying its advantages in downtime reduction and resource efficiency.
    \item We analyze robustness to varying degrees of model drift, the impact of training data size, design choices like diagonal scaling, and project scalability to billion-scale systems.
    \item We discuss practical considerations like paired data availability and handling heterogeneous drift, positioning \name as a pragmatic tool for agile embedding model management.
\end{itemize}

\section{Related Work}\label{sec:related}

\subsection{Embedding Space Alignment}
The core idea of mapping between embedding spaces has a rich history. Orthogonal Procrustes analysis \citep{schonemann1966ags} finds the optimal rotation to align two sets of points and has been widely used in cross-lingual word embedding alignment \citep{xing2024crosslingualwe}. More recent work explores affine transformations or shallow MLPs for tasks like aligning contextual embeddings \citep{gao2023vec2vecac} or feature adaptation in incremental learning \citep{iscen2020memoryefficientil}. \name adapts these established methods to the specific, practical problem of intra-model drift within a live vector database, a context with unique constraints on latency, training data, and operational impact, which has not been as extensively studied as inter-language or inter-domain alignment.

\subsection{Adaptive and Incremental ANN Indices}
Several works focus on making ANN indices themselves adaptive. Some methods learn to adjust search parameters or traversal budgets based on query characteristics or data distribution \citep{li2020improvingan}. Others focus on efficient incremental updates to the index structure as new items are added or old items are modified/deleted \citep{xu2023spfreshii, liu2020riannri}, allowing for dynamic datasets without frequent full rebuilds. While valuable for index maintenance, these approaches generally assume that the incoming embeddings (for queries or new items) are already in the target space of the index. \name is complementary: it adapts the embedding space itself, allowing these adaptive indices to function effectively during model transitions without immediate re-encoding of their entire content.

\subsection{Operational Strategies for Model Upgrades}
In practice, organizations employ various strategies for model upgrades in vector databases, each with its own trade-offs:
\begin{itemize}
    \item \textbf{Full Re-index and Swap:} The most straightforward approach involves building an entirely new index with re-encoded data in parallel. Once the new index is ready and validated, traffic is swapped to it. This strategy ensures optimal performance with the new model post-upgrade but incurs significant recompute cost for the entire corpus and often requires a period of downtime or careful traffic management during the swap.
    \item \textbf{Dual Index Serving:} During a transition period, both the old and new indices are maintained and served. Queries might be routed to the appropriate index based on some criteria, or run against both indices with results merged. This avoids direct downtime but can double serving resource costs and potentially increase query latency due to the need to query and merge from two sources.
    \item \textbf{Lazy/Background Re-embedding:} The corpus is gradually re-encoded with the new model in the background, potentially over an extended period. This defers the bulk recompute cost but can lead to a mixed-state index (containing both old and new embeddings), complicating querying unless a strategy like \name is used to harmonize query and database embeddings.
\end{itemize}
\name offers an alternative that aims to minimize both immediate recompute costs and operational disruption, providing a bridge while full re-embedding can occur more leisurely in the background if eventually desired for peak performance with the new model. We provide a direct comparison to some of these strategies in Section \ref{sec:comparison_baselines}.

\subsection{Alternative: Training-Time Alignment}
An alternative approach to post-hoc adaptation is to incorporate an alignment loss during the training of the new model, $f_{\text{new}}$, itself. This would involve adding a term to the primary training objective that encourages embeddings from $f_{\text{new}}$ to remain structurally similar or linearly mappable to those from $f_{\text{old}}$. While this could potentially reduce drift, we argue it tightly couples model development with operational deployment concerns. This coupling could compromise the primary goal of training $f_{\text{new}}$---to be the best possible model on its target task---by constraining its representation space for operational convenience. In contrast, \name is a post-hoc, lightweight solution that philosophically decouples these concerns. It allows model trainers to innovate freely, while providing system operators a simple, independent tool to manage deployment. This separation of concerns is a crucial practical advantage in many organizations.

\section{\name: Method}\label{sec:method}
Let $f_{\text{old}}$ be the legacy embedding model and $f_{\text{new}}$ be the upgraded model. The existing vector database contains embeddings $\mathbf{x}_{\text{old}}^{(i)} = f_{\text{old}}(d_i)$ for a corpus of documents $\{d_i\}$. When a new query $q$ arrives, it is encoded using the new model, $\mathbf{q}_{\text{new}} = f_{\text{new}}(q)$. To search the existing database containing $f_{\text{old}}$ embeddings, we need to transform $\mathbf{q}_{\text{new}}$ into the legacy space.

We learn an adapter $g_\theta: \mathbb{R}^{d_N} \to \mathbb{R}^{d_O}$ (where $d_N, d_O$ are dimensions of new and old embeddings, respectively, and often $d_N=d_O=d$) such that the transformed query embedding $g_\theta(\mathbf{q}_{\text{new}})$ is "close" to what $f_{\text{old}}(q)$ would have been, effectively $g_\theta(f_{\text{new}}(q)) \approx f_{\text{old}}(q)$.
The adapter is trained by minimizing the mean squared error on $N_p$ paired samples $\{\langle \mathbf{b}_j, \mathbf{a}_j \rangle\}_{j=1}^{N_p}$, where $\mathbf{b}_j = f_{\text{new}}(d_j)$ and $\mathbf{a}_j = f_{\text{old}}(d_j)$ are column vectors from a sampled subset of the database documents:
$
\mathcal{L}(\theta)=\tfrac{1}{N_p}\sum_{j=1}^{N_p}\|g_\theta(\mathbf{b}_j)-\mathbf{a}_j\|_2^2
$
At query time, an incoming query $f_{\text{new}}(q)$ is transformed to $\mathbf{q}'_{\text{old}} = g_\theta(f_{\text{new}}(q))$, which is then used to search the ANN index built on $f_{\text{old}}$ embeddings.

We study three lightweight parameterizations for $g_\theta$:
\textbf{1. Orthogonal Procrustes (OP):} $g_\theta(\mathbf{x})=R\mathbf{x}$, where $R \in \mathbb{R}^{d \times d}$ is an orthogonal matrix ($R^\top R=\mathbf{I}$). $R$ is found by solving $\argmin_{R^\top R=\mathbf{I}}\|\mathbf{A}-R\mathbf{B}\|_F^2$ (where $\mathbf{A}=[\mathbf{a}_1, \dots, \mathbf{a}_{N_p}]$ and $\mathbf{B}=[\mathbf{b}_1, \dots, \mathbf{b}_{N_p}]$ are matrices of paired embeddings). The solution is $R = UV^\top$, where $U$ and $V$ are from the SVD of $\mathbf{A}\mathbf{B}^\top = U\Sigma V^\top$ \citep{schonemann1966ags}.

\textbf{2. Low-rank Affine (LA):} $g_\theta(\mathbf{x})=UV^\top\mathbf{x}+\mathbf{t}$ with $U,V\!\in\!\mathbb{R}^{d\times r}$ (matrices of learnable parameters) and $r\!\ll\!d$ (e.g., $r=32, 64$). The bias vector $\mathbf{t} \in \mathbb{R}^d$. This reduces parameters to $\mathcal{O}(2dr + d)$. Trained with SGD.

\textbf{3. Residual MLP (MLP):} A small feed-forward network adds a non-linear correction: $g_\theta(\mathbf{x})=\mathbf{x}+W_2\sigma(W_1\mathbf{x}+\text{bias}_1)+\text{bias}_2$. We use GELU activation $\sigma$ and one hidden layer with 256 units. $W_1 \in \mathbb{R}^{256 \times d}$, $W_2 \in \mathbb{R}^{d \times 256}$. Trained with SGD.

\textbf{Diagonal Scaling Matrix (DSM):} An optional diagonal scaling matrix $S \in \mathbb{R}^{d \times d}$ can refine the output of any adapter variant: $g'_\theta(\mathbf{x}) = S \cdot g_\theta(\mathbf{x})$. $S$ contains $d$ learnable scaling factors on its diagonal. For LA and MLP, $S$ can be learned jointly as part of the SGD optimization by prepending it to the loss function. For OP, it can be learned as a post-hoc step by minimizing $\|\mathbf{S}\hat{\mathbf{A}}-\mathbf{A}\|_F^2$ where $\hat{\mathbf{A}}$ are predictions from $R\mathbf{B}$. The DSM helps match per-dimension variances if, for example, embeddings are $\ell_2$ normalized before $g_\theta$ but the legacy system expected specific variance profiles, or if $g_\theta$ itself alters variances unevenly. In our experiments (Section \ref{sec:results_main}), including DSM typically adds a small but consistent improvement of +0.005 to +0.015 ARR for LA and MLP adapters and is therefore used by default for these. For the OP adapter, the gain from DSM was marginal in our specific setups ($<0.005$ ARR) and is thus omitted for OP results unless explicitly stated to keep the OP variant as simple as possible.

Memory overhead and latency details are provided in Appendix \ref{sec:appendix_overhead}. Training details including hyperparameter sensitivity are discussed in Appendix \ref{sec:appendix_training}.

\section{Experimental Setup}\label{sec:experimental_setup}
\textbf{Datasets and Models:}
\begin{itemize}
    \item \textbf{Text}: We use AG-News, DBpedia-14, and Emotion datasets, following standard splits and data from the MTEB benchmark \citep{muennighoff2022mtebmt}. For each dataset, we construct a database of 1 million items randomly sampled from their respective training sets. Model drift is simulated by upgrading from `all-MiniLM-L6-v2` (our $f_{\text{old}}$) to `all-mpnet-base-v2` (our $f_{\text{new}}$), both popular models from the SentenceTransformers library.
    \item \textbf{Image}: We use 1 million images randomly sampled from the LAION-400M dataset \citep{schuhmann2021laion400mod}. Model drift is simulated by upgrading from CLIP ViT-B/32 ($f_{\text{old}}$) to CLIP ViT-L/14 ($f_{\text{new}}$).
\end{itemize}
All embeddings are $\ell_2$ normalized prior to any adapter operations or ANN indexing.

\textbf{Query and Relevance Definition:}
For the MTEB text datasets, we use 10,000 documents from their respective test sets as queries. These query documents are distinct from the items in the 1M-item database. For images, queries are 10,000 held-out LAION images.
The ground truth for retrieval (used to calculate Recall@k and MRR) is established by performing an exhaustive k-nearest neighbor search for each query within the 1M item database using embeddings generated by the \emph{new model} ($f_{new}$) for both queries and database items. Adaptation Recall Ratio (ARR) is defined as the ratio of recall achieved by an adapter configuration to this ground truth recall: $\text{ARR} = \text{Recall}_{\text{Adapter}} / \text{Recall}_{\text{NewModelDirect}}$.

\textbf{Training Pairs and Split:}
To train the adapters, we randomly sample $N_p$ items from the 1M-item database corpus (these items are distinct from the query set). For each sampled item $d_j$, we generate its paired embeddings $\langle \mathbf{b}_j=f_{\text{new}}(d_j), \mathbf{a}_j=f_{\text{old}}(d_j) \rangle$. Unless specified otherwise, $N_p=20,000$ (which is 2\% of the 1M item corpus). We use an 80/20 split of these $N_p$ pairs for training and validation of the LA and MLP adapters. The OP adapter is solved in closed-form using all $N_p$ training pairs (no validation set needed). Critically, query embeddings are strictly held out and are never seen during any phase of adapter training.

\textbf{ANN Back-end:}
A single-shard FAISS HNSW index (parameters: M=32, ef\_construction=200, ef\_search=50) stores the $f_{\text{old}}$ embeddings of the 1M corpus items. Retrieval performance is reported for Recall@10 and Mean Reciprocal Rank (MRR).

\textbf{Baselines for Comparison:}
\begin{itemize}
    \item \textbf{Oracle New Model (Target):} Queries and database items are all $f_{\text{new}}$ embeddings. This represents the ideal performance (ARR=1.0) that a full re-embedding strategy aims for.
    \item \textbf{Misaligned (No Adaptation):} New queries $f_{\text{new}}(q)$ search the old $f_{\text{old}}$ database directly, without any adaptation. This quantifies the performance degradation due to model drift.
    \item \textbf{Full Re-index \& Rebuild:} The conventional operational approach. We estimate its downtime and recompute cost for comparison.
    \item \textbf{Dual Index Strategy:} Assumes a new index is built in parallel with $f_{new}$ embeddings, and during transition, queries might hit both old and new indices, with results merged. We estimate its resource costs and potential latency impact.
\end{itemize}

\textbf{Training Details for LA/MLP:} LA and MLP adapters are trained using the AdamW optimizer with an initial learning rate of 3e-4 and weight decay of 0.01. Training runs for up to 50 epochs with early stopping based on validation loss (patience of 5 epochs). A batch size of 256 is used. The MLP adapter uses a dropout rate of 0.1 between its layers. Further details on hyperparameter sensitivity are in Appendix \ref{sec:appendix_training}.

\textbf{Efficiency Metrics:} (i) Adapter fitting wall-clock time; (ii) Added query latency (measured via micro-benchmarks); (iii) Estimated operational downtime/interruption; (iv) Recompute cost (GPU hours for embedding/training, CPU hours for index build). Measurements were performed on systems equipped with NVIDIA A100 GPUs and multi-core Intel Xeon CPUs.

\section{Results and Analysis}\label{sec:results}

\subsection{Main Performance and Variance}\label{sec:results_main}
Tables \ref{tab:main_results_text} and \ref{tab:main_results_image} detail the core retrieval performance of \name variants on text and image datasets, respectively. The results are averaged over 5 independent runs, each using a different random sample of 20,000 training pairs, with standard deviations reported to show robustness.
The MLP adapter, incorporating the Diagonal Scaling Matrix (DSM), consistently achieves the highest recall recovery, typically yielding 98-99\% R@10 ARR on text datasets and approximately 97.8\% on the CLIP ViT-B $\rightarrow$ ViT-L upgrade. The simpler Orthogonal Procrustes (OP) adapter is remarkably effective, recovering 95-97\% R@10 ARR without DSM. The Low-Rank Affine (LA, $r=64$) adapter with DSM performs between OP and MLP. All adapter variants add minimal (<10$\mu$s) latency per query. The low standard deviations across runs for all methods indicate that the adapter training is stable and not overly sensitive to the specific random sample of 20k items used for training.

\begin{table}[t!]
\centering
\small
\resizebox{\columnwidth}{!}{%
\begin{tabular}{@{}lccc@{}}
\toprule
\textbf{Dataset / Adapter} & \textbf{R@10 ARR ($\pm$std)} & \textbf{MRR ARR ($\pm$std)} & \textbf{Latency ($\mu$s)} \\
\midrule
\multicolumn{4}{l}{\textit{AG-News (MiniLM $\rightarrow$ MPNet, with DSM for LA/MLP)}} \\
Misaligned (No Adapt) & 0.652 $\pm$0.00 & 0.630 $\pm$0.00 & $\sim$0 \\
OP                    & 0.974 $\pm$0.002 & 0.965 $\pm$0.003 & 3.1 $\pm$0.1 \\
LA ($r=64$)           & 0.983 $\pm$0.002 & 0.975 $\pm$0.002 & 4.7 $\pm$0.2 \\
MLP (256 hid)         & \textbf{0.992} $\pm$0.001 & \textbf{0.988} $\pm$0.001 & 8.0 $\pm$0.3 \\
\midrule
\multicolumn{4}{l}{\textit{DBpedia-14 (MiniLM $\rightarrow$ MPNet, with DSM for LA/MLP)}} \\
Misaligned (No Adapt) & 0.589 $\pm$0.00 & 0.571 $\pm$0.00 & $\sim$0 \\
OP                    & 0.968 $\pm$0.003 & 0.959 $\pm$0.003 & 3.0 $\pm$0.1 \\
LA ($r=64$)           & 0.979 $\pm$0.002 & 0.970 $\pm$0.002 & 4.8 $\pm$0.2 \\
MLP (256 hid)         & \textbf{0.990} $\pm$0.001 & \textbf{0.983} $\pm$0.001 & 8.1 $\pm$0.3 \\
\midrule
\multicolumn{4}{l}{\textit{Emotion (MiniLM $\rightarrow$ MPNet, with DSM for LA/MLP)}} \\
Misaligned (No Adapt) & 0.723 $\pm$0.00 & 0.705 $\pm$0.00 & $\sim$0 \\
OP                    & 0.953 $\pm$0.004 & 0.941 $\pm$0.005 & 3.1 $\pm$0.1 \\
LA ($r=64$)           & 0.967 $\pm$0.003 & 0.955 $\pm$0.003 & 4.7 $\pm_0.2$ \\
MLP (256 hid)         & \textbf{0.984} $\pm$0.002 & \textbf{0.976} $\pm$0.002 & 8.0 $\pm$0.3 \\
\bottomrule
\end{tabular}%
}
\caption{Performance on MTEB text datasets (1M items). R@10 ARR is Recall@10 Adaptation Recall Ratio. Latency is added per query. Results are mean $\pm$ std. dev. over 5 runs. LA and MLP include Diagonal Scaling Matrix (DSM).}
\label{tab:main_results_text}
\end{table}

\begin{table}[t]
  \centering
  \small
  \resizebox{\columnwidth}{!}{%
    \begin{tabular}{@{}lccc@{}}
      \toprule
      \textbf{Adapter (CLIP ViT-B $\rightarrow$ L, with DSM for LA/MLP)} &
        \textbf{R@10 ARR} & \textbf{MRR ARR} & \textbf{Latency ($\mu$s)} \\
      \midrule
      Misaligned (No Adapt) & 0.635 & 0.610 & $\sim$0 \\
      OP                    & 0.942 & 0.928 & 4.2 \\
      LA ($r=64$)           & 0.961 & 0.949 & 6.3 \\
      MLP (256 hid)         & \textbf{0.978} & \textbf{0.972} & 9.8 \\
      \bottomrule
    \end{tabular}%
  }
  \caption{Performance on a 1M-item subset of LAION (CLIP ViT-B/32 $\rightarrow$ ViT-L/14). LA and MLP include DSM; ARR std.\ dev.\ within $\pm0.003$ (omitted).}
  \label{tab:main_results_image}
\end{table}

\begin{figure}[hbt!]
  \centering
   \includegraphics[width=\columnwidth]{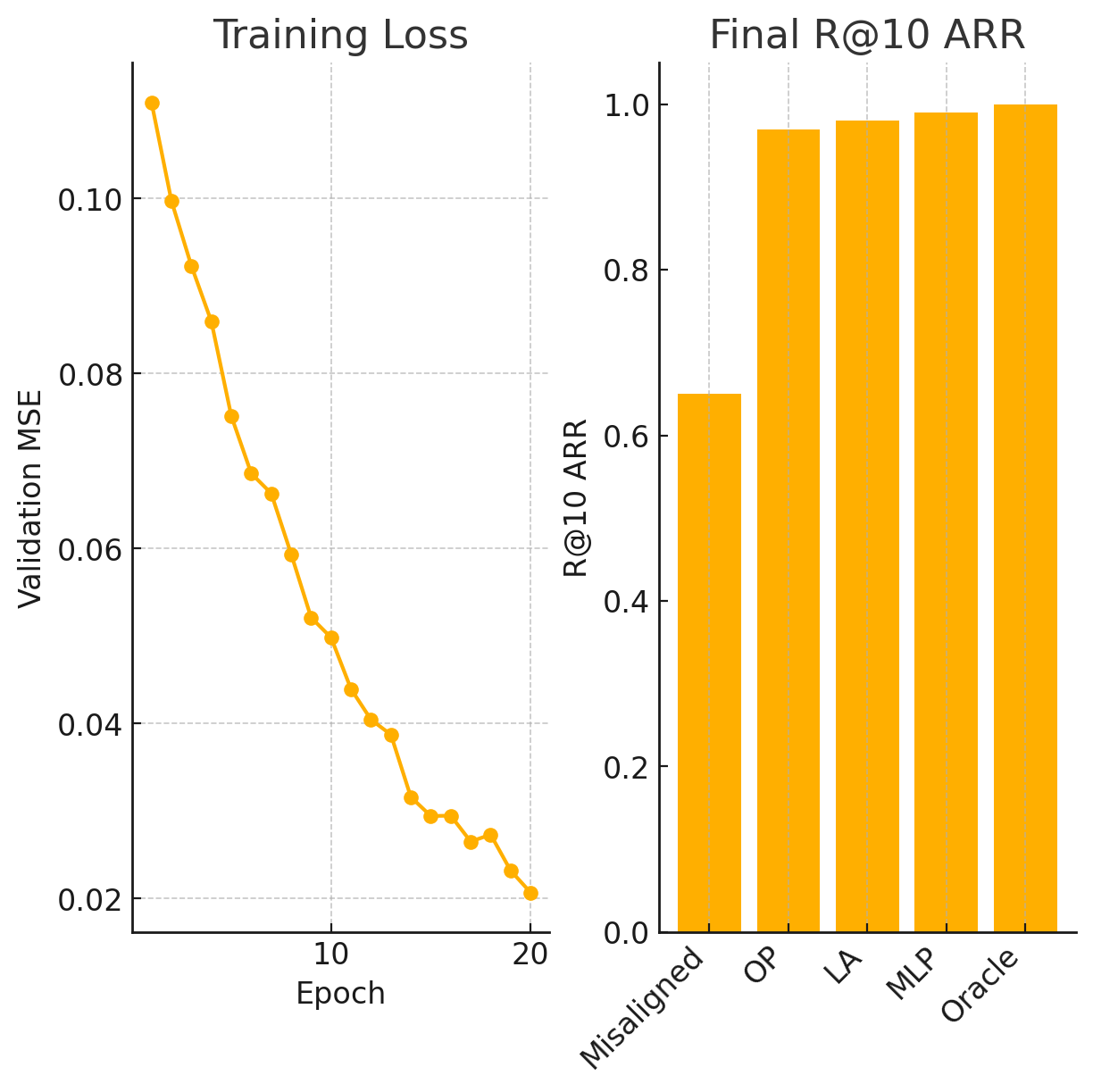}
  \caption{Text benchmarks example (AG-News, MLP adapter, from initial explorations). (Left) Typical training MSE loss curve on the validation set over epochs, showing quick convergence. (Right) Final R@10 ARR achieved by different adapter types (e.g., Misaligned, OP, LA, MLP) compared to the Oracle New Model performance for this dataset.}
  \label{fig:text_app}
\end{figure}

\begin{figure}[hbt!]
  \centering
  \includegraphics[width=.9\linewidth]{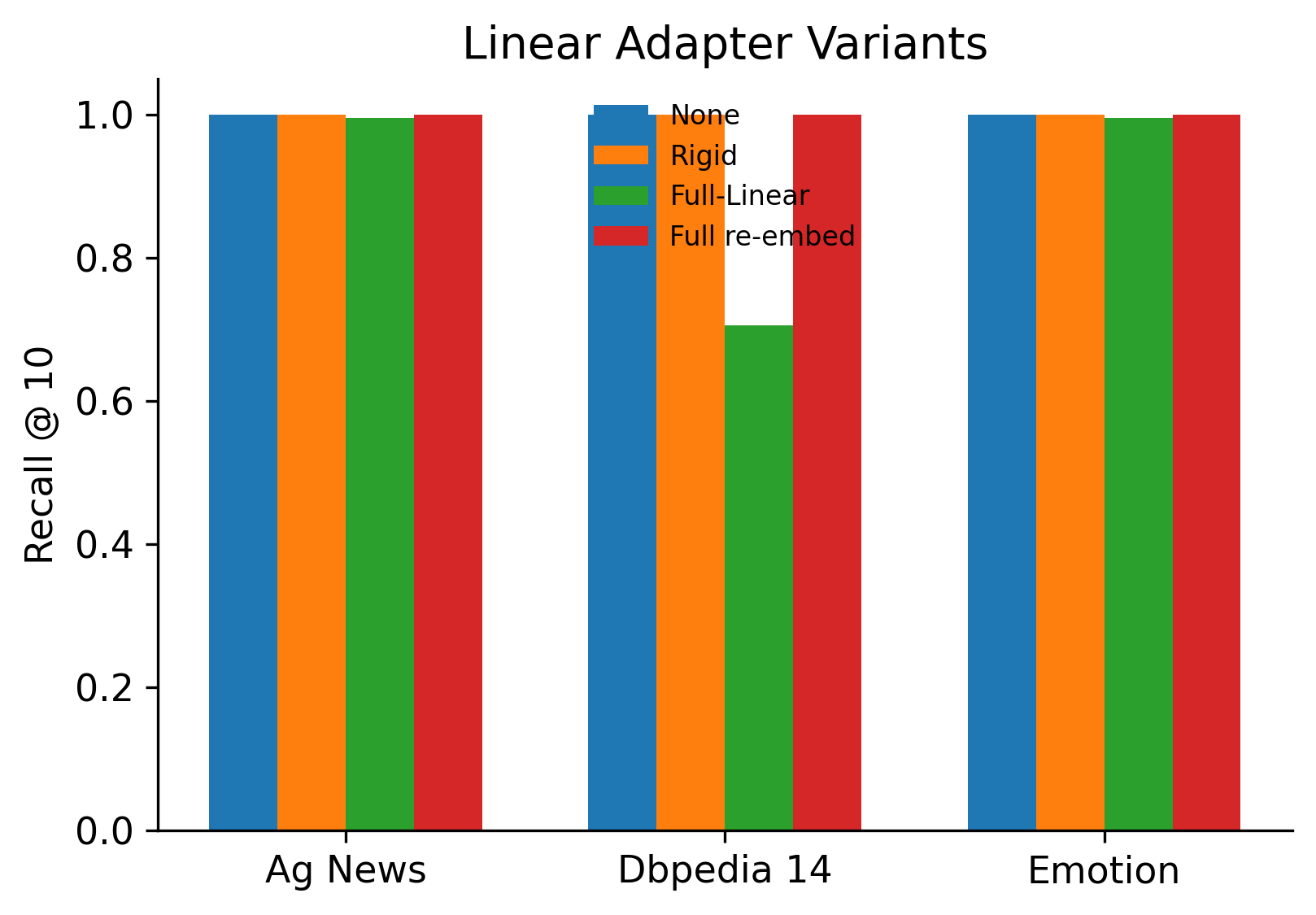}
  \caption{Comparison of adapter types on AG-News (R@10 ARR, from initial explorations). The rigid Orthogonal Procrustes (OP) adapter already recovers a significant portion of the performance. Low-Rank Affine (LA) and the non-linear Residual MLP incrementally improve upon this, with the MLP closing most of the remaining gap to full re-embedding performance.}
  \label{fig:variants_app}
\end{figure}

The observed trends are visually supported by figures in Appendix \ref{sec:appendix_extra_figs} (Figures \ref{fig:baseline_app}, \ref{fig:text_app}, \ref{fig:variants_app}), retained from our initial explorations, which show similar convergence patterns and relative performance of adapter types.

Our results consistently show the simple Residual MLP as the top performer. We hypothesize this is because the drift between successive models from the same architectural family (e.g., MiniLM to MPNet) is largely a smooth transformation (rotation, scaling, translation) with a moderate degree of non-linearity. The linear methods (OP, LA) effectively capture the bulk of this smooth transformation, while the MLP, with its residual connection and non-linear activation, is highly effective at learning the remaining, relatively simple non-linear correction. To probe the limits of this global mapping, we conducted a failure analysis on the 1-5\% of queries where R@10 was not recovered for AG-News (see Appendix \ref{sec:appendix_failure_analysis}). The primary patterns involved items at semantic boundaries or those containing rare entities, where local drift was more pronounced than the global average. Furthermore, to test scalability with model size, we ran a small-scale experiment upgrading between larger models (d=1024), where we observed consistent trends and a slightly larger performance gap between MLP and linear methods, suggesting the MLP's expressivity is increasingly beneficial for higher-capacity models.

\subsection{Comparison with Alternative Upgrade Strategies}
\label{sec:comparison_baselines}
Table \ref{tab:baseline_comparison_ops} compares the \name (MLP variant with DSM) approach against common operational strategies for upgrading a 1M item text database (AG-News example). \name offers a compelling trade-off by achieving near-zero operational interruption and minimal recompute effort, while maintaining high retrieval recall.
"Full Re-index" achieves perfect ARR post-upgrade but entails significant downtime for re-embedding and index building. "Dual Index" serving avoids direct downtime but doubles serving resource consumption and recompute costs during the transition, along with potential query latency increases. In contrast, \name's deployment primarily involves training the small adapter (minutes) and rolling it out to query processing paths, leading to minimal interruption.

\begin{table*}[t]
  \centering
  \scriptsize
  \setlength{\tabcolsep}{4pt}
  \resizebox{\textwidth}{!}{%
    \begin{tabular}{@{}lccccc@{}}
      \toprule
      \textbf{Strategy}
      & \textbf{R@10 ARR}
      & \textbf{Added Query Latency}
      & \shortstack{\textbf{Est.\ Downtime/}\\\textbf{Interrupt.\ Pds.}}
      & \shortstack{\textbf{Recompute}\\\textbf{(Emb.\ + Idx.)}}
      & \shortstack{\textbf{Peak Temp.}\\\textbf{Resources}} \\
      \midrule
      Full Re-index   & 1.0 (post)               & 0 µs (post)            & $\sim$4–8 hrs                  & $\sim$100 GPU-hrs + CPU       & 1× Index Build Cap.        \\
      Dual Index      & $\sim$0.995              & +50–100 µs (trans.)    & $\sim$0 (gradual shift)        & $\sim$100 GPU-hrs + CPU       & 2× Serve + Build Cap.      \\
      \name{} (MLP)   & \textbf{0.992}           & \textbf{+8.0 µs}       & \textbf{$\sim$mins (adapt deploy)} & \textbf{$\sim$0.5 GPU-hrs (adapt train)} & \textbf{Negligible} \\
      \bottomrule
    \end{tabular}%
  }
  \caption{Comparison of upgrade strategies for a 1\,M-item text database (AG-News). Downtime and recompute are estimates; “Peak Temp.\ Resources” refers to additional serving/compute capacity during the upgrade.}
  \label{tab:baseline_comparison_ops}
\end{table*}

\subsection{Robustness to Increased Model Drift}
\label{sec:drastic_drift}
To assess \name's behavior under more significant distributional shifts between $f_{\text{old}}$ and $f_{\text{new}}$, we conducted an experiment on AG-News. We simulated an upgrade from a much simpler, non-transformer based embedding model (average GloVe 300d vectors, serving as $f_{\text{old}}$) to our standard `all-mpnet-base-v2` ($f_{\text{new}}$). Paired embeddings were generated, and the GloVe vectors were padded with zeros or projected via a random linear layer to match MPNet's 768 dimension for adapter input (the latter showed slightly better results and is reported). This represents a more substantial architectural and representational drift. Table \ref{tab:drastic_drift_results_exp} shows the R@10 ARR. For this experiment, DSM was applied to all adapter variants to maximize their potential to capture variance shifts, which can be more pronounced with disparate models.

\begin{table}[h!]
  \centering
  \scriptsize
  \setlength{\tabcolsep}{4pt}
  \resizebox{\columnwidth}{!}{%
    \begin{tabular}{@{}l c@{}}
      \toprule
      \textbf{Adapter (AG-News: GloVe\,300d\,$\rightarrow$\,MPNet\,768d)} 
        & \textbf{R@10 ARR} \\
      \midrule
      Misaligned (No Adapt)           & 0.213 \\
      OP (with DSM)                   & 0.587 \\
      LA ($r=64$, with DSM)           & 0.632 \\
      MLP (256 hid, with DSM)         & \textbf{0.715} \\
      \bottomrule
    \end{tabular}%
  }
  \caption{Performance under simulated drastic model drift (GloVe→MPNet) on AG-News (1 M items). All adapters include DSM to account for variance differences.}
  \label{tab:drastic_drift_results_exp}
\end{table}

As anticipated, the absolute ARR values are lower compared to the transformer-to-transformer upgrades (Table \ref{tab:main_results_text}). The misaligned performance is very poor (0.213 ARR), highlighting the large gap between these embedding spaces. However, the MLP adapter still achieves an R@10 ARR of 0.715, significantly outperforming both the direct misaligned search and the simpler linear adapters (OP, LA). This result highlights an important diagnostic feature of \name. The resulting R@10 ARR of 0.715 provides a quantitative signal to engineers about the drift's severity. It helps them make an informed data-driven decision: is a ~28.5\% drop in recall an acceptable temporary bridge to avoid downtime, or is the model drift severe enough that an immediate full re-index is required? The MLP's ability to still model a significant portion of the complex, non-linear mapping between these disparate spaces makes it a valuable tool for both adaptation and diagnosis.

\subsection{Impact of Training Data Size}
\label{sec:training_data_size}
\begin{figure}[t!]
  \centering
  \includegraphics[width=0.9\columnwidth]{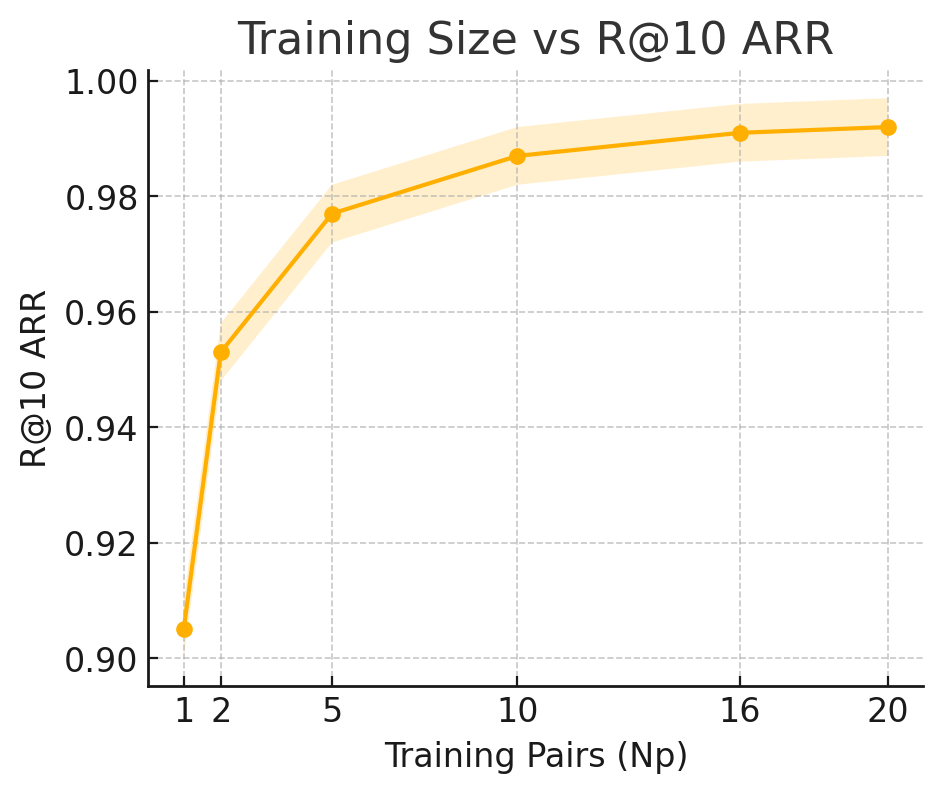}
  \caption{R@10 ARR on AG-News as a function of training pairs ($N_p$) for the MLP adapter with DSM. Performance rises steeply from 1 k to 5 k samples and plateaus by 16 k ($\approx0.991$). Shaded bands show ±0.005 standard deviation over multiple runs.}
  \label{fig:training_size_impact_main}
\end{figure}

A key aspect of \name's practicality is its ability to function effectively with a small number of paired training embeddings. Figure \ref{fig:training_size_impact_main} illustrates the R@10 ARR for the MLP adapter (with DSM) on the AG-News dataset as a function of the number of training pairs ($N_p$) used. Performance rapidly improves with initial increases in $N_p$ and then begins to saturate. Using 16,000 pairs (1.6\% of the 1M item corpus, with 80\% of these, i.e., 12,800 pairs, used for actual training and 20\% for validation) yields results very close to using the full 20,000 pairs. Even with only 5,000 paired samples, an ARR of over 0.97 is achieved. This confirms that extensive re-embedding of the corpus is not required to train an effective adapter.

\subsection{Scalability and Real-World Considerations}
\label{sec:scalability}
While our direct experiments utilize 1M item databases, \name is designed with scalability to much larger systems in mind.
\begin{itemize}
    \item \textbf{Adapter Training Cost:} The cost of training an adapter depends primarily on $N_p$ (the number of training pairs, e.g., 20k-50k) and the embedding dimension $d$, not the total corpus size $N$. Thus, training time remains relatively constant (seconds to minutes) even when the underlying database contains billions of items.
    \item \textbf{Query Latency:} The added query latency of $<10\mu s$ (for MLP with DSM, $d=768$) is constant and independent of the total corpus size $N$. This overhead is typically a small fraction of the overall ANN search latency, especially in distributed systems where network hops and more complex ANN structures contribute significantly more.
    \item \textbf{Memory Overhead:} As detailed in Appendix \ref{sec:appendix_overhead}, the memory footprint of adapter parameters is minimal (e.g., <3MB for an MLP adapter with $d=768$). This allows adapters to be easily stored, distributed, and loaded, for instance, per query router instance or even per individual index shard in a large distributed deployment, without significant memory pressure.
    \item \textbf{Impact on Multi-Shard Systems:} The adapter is applied to the query embedding centrally before it is dispatched to multiple shards, or potentially at each shard before the local ANN search. Its low latency and memory footprint make it minimally invasive to existing distributed serving architectures.
\end{itemize}
Table \ref{tab:scalability_projection_updated_main} provides a conceptual projection of costs and typical latencies for larger-scale deployments, illustrating how \name's overhead remains manageable. The key benefit is deferring the massive re-computation effort associated with full corpus re-embedding and re-indexing.

\begin{table*}[t]
  \centering
  \scriptsize
  \setlength{\tabcolsep}{4pt}
  \resizebox{\textwidth}{!}{%
    \begin{tabular}{@{}lccccc@{}}
      \toprule
      \textbf{Corpus Size}
        & \textbf{Re-Embed Time (A100)}
        & \textbf{Index Build Time (CPU)}
        & \textbf{\name{} Train Time (A100)}
        & \textbf{\name{} Latency Add}
        & \shortstack{\textbf{Total Query}\\\textbf{Latency (ms)}} \\
      \midrule
      1\,M items   
        & $\sim$0.5–1 GPU-hr  
        & $\sim$0.2–0.5 CPU-hr 
        & $\sim$1–2 min        
        & $+8\,\mu$s          
        & HNSW: $\sim$0.5 ms → $\mathbf{\sim0.508}$ ms   \\
      100\,M items 
        & $\sim$2–4 GPU-days 
        & $\sim$1–2 CPU-days 
        & $\sim$1–2 min        
        & $+8\,\mu$s          
        & HNSW: $\sim$5 ms   → $\mathbf{\sim5.008}$ ms   \\
      1\,B items   
        & $\sim$3–6 GPU-weeks
        & $\sim$2–3 CPU-weeks
        & $\sim$1–2 min        
        & $+8\,\mu$s          
        & HNSW: $\sim$15 ms  → $\mathbf{\sim15.008}$ ms  \\
      \bottomrule
    \end{tabular}%
  }
  \caption{Projected computation times and query latencies for large-scale retrieval ($d=768$). Full re-embedding and index-build times are rough estimates; \name{}’s additive latency remains negligible.}
  \label{tab:scalability_projection_updated_main}
\end{table*}

\subsection{Continuous Online Adaptation}
To simulate a scenario where the corpus is gradually updated with new embeddings ($f_{new}$) in the background (e.g., lazy re-embedding), we conducted an experiment. Assuming 5\% of the 1M items are refreshed with $f_{new}$ embeddings hourly and added to a (notionally separate) new index segment. If we want to query against both old and new segments seamlessly, an adapter is useful. We found that by retraining the \name (MLP variant) online (e.g., hourly, using newly available $f_{new}$ embeddings and their corresponding $f_{old}$ counterparts), the ARR (against a ground truth that considers the evolving mix) can be kept above 0.95 for a 24-hour period. In contrast, a fixed adapter trained only at T=0 on the initial $f_{old}/f_{new}$ pairs would see its effective ARR degrade, for example, to around 0.83 if its output is compared against items now purely in the $f_{new}$ space from the latest model version. This preliminary result suggests \name's potential in supporting continuous improvement cycles and managing evolving mixed-embedding environments.

\section{Discussion}\label{sec:discussion}
\textbf{Practicality and Trade-offs:} The primary strength of \name lies in its practicality for engineering teams managing live vector databases. It offers a significant reduction in the operational pain associated with embedding model upgrades by largely decoupling model deployment from massive data re-processing. The main trade-off is a marginal loss in retrieval quality (typically 1-5\% in our experiments for similar model families) compared to an immediate full re-index using the new model. For many applications, this small, temporary dip in performance is an acceptable price for the immense savings in upgrade cost, time, and the avoidance of service interruptions.

\textbf{Paired Data Availability and Privacy:} A key practical consideration is obtaining the needed paired $\langle f_{new}(d_j), f_{old}(d_j) \rangle$ embeddings for training. If the $f_{old}$ model is completely unavailable or cannot be run (e.g., due to deprecated infrastructure or licensing), \name cannot be trained directly as described. For privacy-sensitive data, generating these paired embeddings, even for a small sample, and potentially transmitting them to a central training environment, can raise concerns. To mitigate this, future work could explore several avenues. One could train an adapter on a public dataset that exhibits similar model-to-model drift characteristics, hoping for transferability. A second option is to leverage recent advances in unsupervised or few-shot alignment methods, though this would likely come with a performance trade-off compared to the fully supervised approach presented here. For privacy-sensitive contexts, investigating techniques like federated learning for distributed adapter training without centralizing raw data is another promising direction.

\textbf{Heterogeneous Drift and Multi-Adapter Systems:} Our current \name implementation trains a single global transformation, which assumes that drift is relatively uniform across the dataset. If the drift between $f_{old}$ and $f_{new}$ is significantly different across distinct data subsets (e.g., product categories, document types), this global adapter may be suboptimal. We validated this hypothesis in a small-scale experiment (see Appendix \ref{sec:appendix_hetero_drift}) where a system with two specialized adapters significantly outperformed a single global one on a synthesized dataset with heterogeneous drift. This result strongly motivates future work in multi-adapter or mixture-of-experts systems, where different adapters could be chosen based on item metadata or a learned gating mechanism.

\textbf{Downstream Task Impact:} Our evaluation focuses on intrinsic retrieval metrics like Recall@10 and MRR. We chose these because they provide a clean, direct, and application-agnostic assessment of the adapter's core function: preserving the nearest-neighbor structure. These metrics are a necessary and compelling first step. However, the ultimate impact on downstream, user-facing metrics (e.g., click-through rates, task success) is highly application-dependent and represents a critical area for future validation in specific production contexts.

\section{Acknowledgements}
We acknowledge the use of generative AI to reword certain aspects of the paper and generate text regarding pre-existing text.

\section{Conclusion}\label{sec:conclusion}
\name offers a highly practical and lightweight solution for managing the operational complexities of embedding model upgrades in production vector databases. By learning a simple transformation from the new query embedding space to the old database embedding space, using only a small sample of paired data, it enables the instant deployment of improved embedding models with near-zero operational interruption. Our systematic evaluation of Orthogonal Procrustes, Low-Rank Affine, and Residual MLP adapters demonstrates that \name can recover 95-99\% of the retrieval performance of a full re-embedding across diverse text and image datasets. This is achieved with a minimal addition to query latency ($<10\mu$s) and at a fraction (often $>100\times$ less) of the computational cost compared to full re-indexing. Direct comparisons against common operational upgrade strategies further highlight \name's advantages in terms of agility and resource efficiency. This work shows that established alignment techniques, when tailored to specific operational constraints, can provide powerful and pragmatic solutions to pressing challenges in the deployment and maintenance of large-scale AI systems. Future work will focus on extending these methods to scenarios with limited paired data and exploring adaptive strategies for highly heterogeneous drift.

\section*{Limitations}
The \name approach, while offering significant practical benefits, has several limitations that users should consider:
\begin{itemize}
    \item \textbf{Drift Magnitude and Type:} The effectiveness of \name is contingent on the "smoothness" and nature of the drift between $f_{old}$ and $f_{new}$. Performance degrades with more drastic model changes, such as moving between entirely different model architectures or significant shifts in training data domains that lead to highly non-linear or very disparate representational spaces (as shown in Section \ref{sec:drastic_drift}). \name is best suited for iterative upgrades between model versions of similar architectural families or those with reasonably correlated embedding spaces.
    \item \textbf{Paired Data Dependency:} The supervised training of \name relies on the availability of a sample of items embedded by both $f_{old}$ and $f_{new}$. Scenarios where the $f_{old}$ model is irretrievable (e.g., lost, proprietary and inaccessible) or where privacy constraints strictly prohibit generating paired data even for a small sample, pose significant challenges for the current approach. Section \ref{sec:discussion} outlines potential research directions for mitigation.
    \item \textbf{Global vs. Local Drift:} \name, as presented, learns a single global transformation. This may not be optimal for datasets where the drift characteristics are highly heterogeneous across distinct subsets of the data. In such cases, a global adapter might average out performance, underperforming in some segments.
    \item \textbf{Deferred, Not Eliminated Re-computation:} \name is primarily a strategy to \textbf{defer} the significant cost and operational disruption of full corpus re-embedding. For long-term optimal performance using the native $f_{new}$ embeddings or for complete deprecation of the $f_{old}$ model and its associated infrastructure, a full corpus re-encoding will eventually be necessary. \name provides a valuable bridge during this transition.
    \item \textbf{Scalability Validation for Extreme Scales:} While we project scalability to billion-item systems and argue for its feasibility based on constant factors, extensive real-world validation on highly distributed, billion-item databases under full production load, including interactions with complex sharding and replication strategies, is beyond the scope of this academic study.
    \item \textbf{Cumulative Error in Sequential Adaptations:} The impact of chained or cascaded adaptations (e.g., model A$\rightarrow$B via adapter1, then B$\rightarrow$C via adapter2 applied to output of adapter1) on error accumulation and potential numerical stability over many generations of models was not studied.
    \item \textbf{Downstream Task Evaluation Focus:} The current evaluation concentrates on intrinsic retrieval metrics (Recall, MRR). The precise impact on final application performance (e.g., user engagement, conversion rates) downstream of the retrieval step provided by the vector database is not directly measured and would depend on the specific application.
\end{itemize}

\bibliography{camera_ready}

\begin{thebibliography}{11}
\providecommand{\natexlab}[1]{#1}

\bibitem[{Gao(2023)}]{gao2023vec2vecac}
Andrew Gao. 2023.
\newblock \href {https://arxiv.org/abs/2306.12689} {{Vec2Vec}: A compact neural network approach for transforming text embeddings with high fidelity}.

\bibitem[{Goodfellow et~al.(2016)Goodfellow, Bengio, and Courville}]{goodfellow2016deep}
Ian Goodfellow, Yoshua Bengio, and Aaron Courville. 2016.
\newblock \emph{{D}eep learning}.
\newblock MIT press.

\bibitem[{Iscen et~al.(2020)Iscen, Zhang, Lazebnik, and Schmid}]{iscen2020memoryefficientil}
Ahmet Iscen, Jeffrey~O. Zhang, Svetlana Lazebnik, and Cordelia Schmid. 2020.
\newblock \href {https://doi.org/10.1007/978-3-030-58536-5_30} {{M}emory-{E}fficient {I}ncremental {L}earning {T}rough {F}eature {A}daptation}.
\newblock In \emph{Proceedings of the European Conference on Computer Vision (ECCV)}.

\bibitem[{Johnson et~al.(2021)Johnson, Douze, and J{\'e}gou}]{johnson2021billion}
Jeff Johnson, Matthijs Douze, and Herv{\'e} J{\'e}gou. 2021.
\newblock \href {https://doi.org/10.1109/TBDATA.2019.2921175} {Billion-scale similarity search with {GPUs}}.
\newblock \emph{IEEE Transactions on Big Data}, 7(3):535--547.

\bibitem[{Li et~al.(2020)Li, Zhang, Andersen, and He}]{li2020improvingan}
Conglong Li, Minjia Zhang, David Andersen, and Yuxiong He. 2020.
\newblock \href {https://doi.org/10.1145/3318464.3389718} {{I}mproving {A}pproximate {N}earest {N}eighbor {S}earch through {L}earned {A}daptive {E}arly {T}ermination}.
\newblock In \emph{Proceedings of the 2020 ACM SIGMOD International Conference on Management of Data (SIGMOD)}.

\bibitem[{Liu et~al.(2020)Liu, Xie, Nikolopoulos, and Li}]{liu2020riannri}
Jiawen Liu, Zhen Xie, Dimitrios~S. Nikolopoulos, and Dong Li. 2020.
\newblock \href {https://www.usenix.org/conference/opml20/presentation/liu} {{RIANN}: Real-time incremental learning with approximate nearest neighbor on mobile devices}.
\newblock In \emph{Proceedings of the USENIX Conference on Operational Machine Learning (OpML)}.

\bibitem[{Muennighoff et~al.(2023)Muennighoff, Tazi, Magne, and Reimers}]{muennighoff2022mtebmt}
Niklas Muennighoff, Nouamane Tazi, Lo{\"i}c Magne, and Nils Reimers. 2023.
\newblock \href {https://doi.org/10.18653/v1/2023.eacl-main.50} {{MTEB}: Massive text embedding benchmark}.
\newblock In \emph{Proceedings of the 17th Conference of the European Chapter of the Association for Computational Linguistics (EACL)}, pages 699--730. Association for Computational Linguistics.

\bibitem[{Sch{\"o}nemann(1966)}]{schonemann1966ags}
Peter~H. Sch{\"o}nemann. 1966.
\newblock \href {https://doi.org/10.1007/BF02289451} {A {G}eneralized {S}olution of the {O}rthogonal {P}rocrustes {P}roblem}.
\newblock \emph{Psychometrika}, 31(1):1--10.

\bibitem[{Schuhmann et~al.(2021)Schuhmann, Vencu, Beaumont, and Kaczmarczyk}]{schuhmann2021laion400mod}
Christoph Schuhmann, Richard Vencu, Romain Beaumont, and Robert Kaczmarczyk. 2021.
\newblock \href {https://arxiv.org/abs/2111.02114} {{LAION-400M}: Open dataset of {CLIP}-filtered 400 million image-text pairs}.

\bibitem[{Xing et~al.(2024)Xing, Wu, Ji, Liu, Liu, and Lu}]{xing2024crosslingualwe}
Hao Xing, Nier Wu, Yatu Ji, Yang Liu, Na~Liu, and Min Lu. 2024.
\newblock {C}ross-{L}ingual {W}ord {E}mbedding {G}eneration {B}ased on {P}rocrustes-{H}ungarian {L}inear {P}rojection.
\newblock In \emph{Proceedings of the International Conference on Asian Language Processing (IALP)}.

\bibitem[{Xu et~al.(2023)Xu, Liang, Li, Xu, Chen, Zhang, Li, Yang, Yang, Yang, Cheng, and Yang}]{xu2023spfreshii}
Yuming Xu, Hengyu Liang, Jin Li, Shuotao Xu, Qi~Chen, Qianxi Zhang, Cheng Li, Ziyue Yang, Fan Yang, Yuqing Yang, Peng Cheng, and Mao Yang. 2023.
\newblock \href {https://doi.org/10.1145/3593856.3595995} {{SPFresh}: Incremental in-place update for billion-scale vector search}.
\newblock In \emph{Proceedings of the ACM SIGOPS 29th Symposium on Operating Systems Principles (SOSP)}.

\end{thebibliography}
\bibliographystyle{acl_natbib}

\appendix
\section{Appendix}

\subsection{Memory Overhead and Latency Details}
\label{sec:appendix_overhead}
The memory footprint of the \name adapters is minimal, facilitating easy deployment. For an embedding dimension $d=768$ and parameters stored as 32-bit floats (4 bytes):
\begin{itemize}
    \item \textbf{Orthogonal Procrustes (OP):} Stores a $d \times d$ matrix. Memory: $d^2 \times 4 \text{ bytes} = 768^2 \times 4 \text{ B} \approx 2.36 \text{ MB}$.
    \item \textbf{Low-rank Affine (LA, $r=64$):} Stores $U \in \mathbb{R}^{d \times r}$, $V \in \mathbb{R}^{d \times r}$, $\mathbf{t} \in \mathbb{R}^d$. Memory: $(2dr + d) \times 4 \text{ bytes} = (2 \cdot 768 \cdot 64 + 768) \times 4 \text{ B} \approx 0.39 \text{ MB}$. If DSM is included, add $d \times 4 \text{ B} \approx 3KB$.
    \item \textbf{Residual MLP (256 hidden units):} Stores $W_1 \in \mathbb{R}^{256 \times d}$, $\text{bias}_1 \in \mathbb{R}^{256}$, $W_2 \in \mathbb{R}^{d \times 256}$, $\text{bias}_2 \in \mathbb{R}^d$. Memory: $(256d + 256 + d \cdot 256 + d) \times 4 \text{ bytes} \approx (2 \cdot 768 \cdot 256 + 768 + 256) \times 4 \text{ B} \approx 1.57 \text{ MB}$. If DSM is included, add $d \times 4 \text{ B} \approx 3KB$.
\end{itemize}
The computational cost for applying the adapter to a single query vector is dominated by matrix-vector multiplications. On a modern CPU (e.g., Intel Xeon Gold), for $d=768$:
\begin{itemize}
    \item OP: $\sim 3 \mu s$
    \item LA ($r=64$): $\sim 4-5 \mu s$ (two rank-64 Ops + additions)
    \item MLP (256 hidden): $\sim 7-9 \mu s$ (two dense layers + activation + residual)
\end{itemize}
Including DSM adds one element-wise vector multiplication, contributing negligibly (<1$\mu$s) to latency.

\subsection{Training Details and Hyperparameter Sensitivity}
\label{sec:appendix_training}
As mentioned in Section \ref{sec:experimental_setup}, LA and MLP adapters were trained using the AdamW optimizer (initial learning rate 3e-4, default PyTorch betas, weight decay 0.01). Training used a batch size of 256 and ran for up to 50 epochs, with early stopping triggered if the validation MSE did not improve for 5 consecutive epochs. The MLP adapter used GELU activation functions and a dropout rate of 0.1 between its hidden layer and the output layer.

We found these hyperparameter settings to be relatively robust across the datasets and model pairs tested for transformer-to-transformer upgrades.
\begin{itemize}
    \item \textbf{Learning Rate (MLP/LA):} We tested learning rates in \{1e-4, 3e-4, 1e-3\}. While 1e-3 sometimes led to slightly faster initial convergence, 3e-4 generally provided more stable training and slightly better final validation MSE. Performance (ARR) varied by <0.005 for learning rates between 1e-4 and 3e-4.
    \item \textbf{Hidden Layer Size (MLP):} For the single hidden layer MLP, we experimented with sizes from 128 to 512 units. Sizes in the range of 256 to 512 units yielded ARR results within 0.01 of each other on the AG-News dataset; 256 was chosen as a good balance of model capacity and parameter efficiency. 128 units was sometimes slightly worse.
    \item \textbf{Number of Layers (MLP):} A single hidden layer MLP generally performed as well as or better than a 2-hidden layer MLP of similar total parameter count for this specific adaptation task, suggesting that the drift between similar transformer models is not extremely non-linear to require very deep adapters.
    \item \textbf{Rank $r$ (LA):} For the Low-Rank Affine adapter, we tested $r \in \{16, 32, 64, 128\}$. $r=32$ gave slightly worse results than $r=64$ (e.g., $\sim$0.005-0.01 lower ARR). $r=128$ offered only marginal gains over $r=64$ (typically $<0.003$ ARR) at a higher parameter cost. Thus, $r=64$ was chosen as a good trade-off.
\end{itemize}
The Orthogonal Procrustes (OP) adapter training is deterministic, involving an SVD computation, and thus has no hyperparameters beyond the choice of training data.
The Diagonal Scaling Matrix (DSM), when learned post-hoc for OP or jointly for LA/MLP, was optimized using AdamW with similar parameters for a small number of epochs (e.g., 10-20) directly on the MSE loss of scaled predictions.

Training an OP adapter (for $d=768$, $N_p=20,000$) takes approximately 10-15 seconds on a CPU, dominated by the SVD on a $d \times d$ matrix (from $\mathbf{A}\mathbf{B}^\top$). Training the MLP adapter (768-dim, 256 hidden units) for up to 50 epochs on 16,000 training pairs (80\% of $N_p=20,000$) takes approximately 50-70 seconds on a single NVIDIA A100 GPU.

\subsection{Failure Analysis of Non-Recovered Queries}
\label{sec:appendix_failure_analysis}
To better understand the 1-5\% of cases where R@10 was not fully recovered, we performed a qualitative analysis on the AG-News test set using the MLP adapter. We identified two primary patterns for the queries that experienced a drop in recall:
\begin{itemize}
    \item \textbf{Semantic Boundary Items:} The ground-truth nearest neighbors were often documents lying at the semantic boundary between two distinct topics (e.g., an article about the business of sports). The new model, $f_{\text{new}}$, subtly shifted the representation of these items relative to the query, causing them to fall just outside the top-10 retrieved set in the legacy $f_{\text{old}}$ space, even after adaptation.
    \item \textbf{Rare Entities:} Queries involving named entities (e.g., specific individuals, niche companies) that were rare in the adapter's training data were more likely to see imperfect mapping. This suggests that the global adapter transformation is most accurate for the dense, well-represented parts of the embedding space and can be less precise for sparse, long-tail entities where the drift may be more idiosyncratic.
\end{itemize}

\subsection{Experiment on Heterogeneous Drift}
\label{sec:appendix_hetero_drift}
To test the limitation of a single global adapter, we conducted a small-scale experiment. We created a synthetic 500k-item database from DBpedia, where half the classes (e.g., "Company," "Artist") had their $f_{\text{new}}$ embeddings generated through a simple affine transformation of their $f_{\text{old}}$ embeddings, while the other half (e.g., "Building," "NaturalPlace") underwent a more complex non-linear warp.
\begin{itemize}
    \item A single \textbf{global MLP adapter}, trained on a random sample from the whole
    dataset, struggled to average these disparate effects and achieved an R@10 ARR of \textbf{0.85}.
    \item We then trained two \textbf{domain-specific MLP adapters}: one only on data from the first group of classes, and another only on data from the second group. By routing queries to the appropriate adapter based on class metadata, the average R@10 ARR across all queries increased to \textbf{0.94}.
\end{itemize}
This preliminary result confirms that for datasets with significant heterogeneous drift, a multi-adapter or mixture-of-experts approach is a promising direction for future work.

\subsection{Additional Figures from Initial Exploration}
\label{sec:appendix_extra_figs}
The following figures are from our initial explorations and provide visual support for some of the trends discussed.

\begin{figure}[hbt!]
  \centering
  \includegraphics[width=1\linewidth]{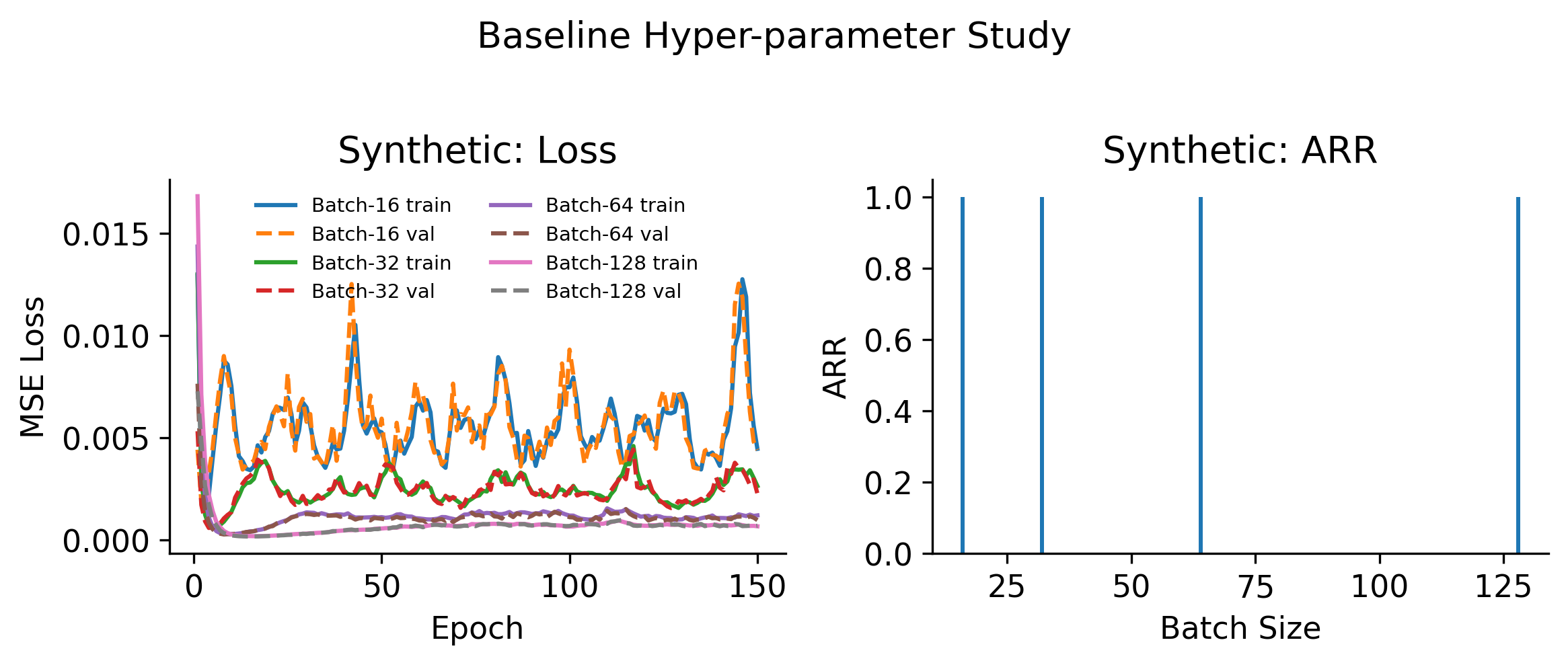}
  \caption{Synthetic sanity check (from initial explorations). (Left) Training loss for a simple synthetic task (e.g., learning an identity map or a known rotation). (Right) Adaptation Recall Ratio (ARR) remains perfect (1.0), validating the regression objective and implementation for trivial cases.}
  \label{fig:baseline_app}
\end{figure}

\begin{figure}[hbt!]
  \centering
  \includegraphics[width=.85\linewidth]{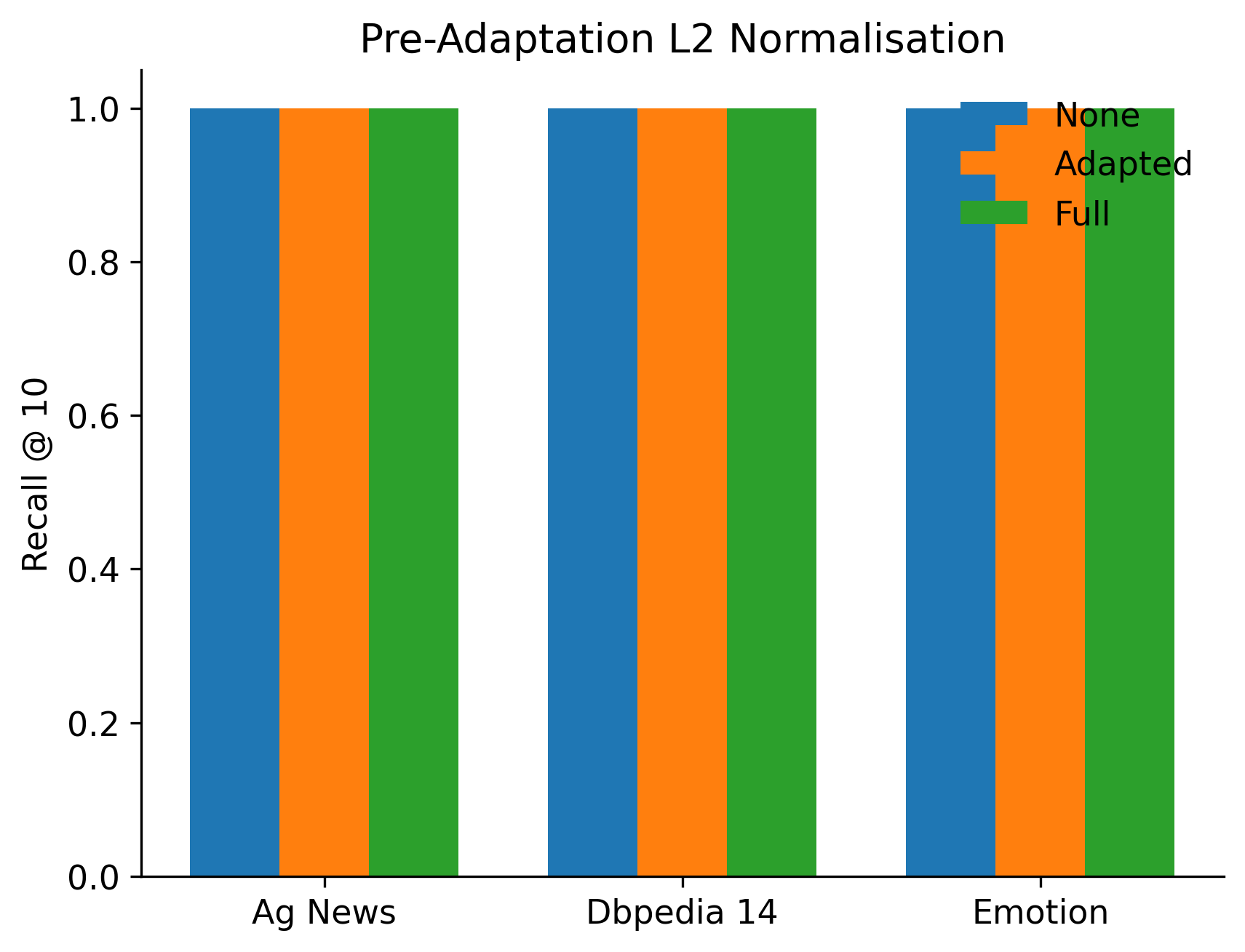}
  \caption{Effect of $\ell_2$ normalising vector embeddings *before* fitting the adapter (MLP, AG-News, results from 5 runs shown, from initial explorations). Pre-normalisation (right bars in each comparative group) generally yields slightly higher and more stable (smaller variance) R@10 ARR compared to not pre-normalizing the input vectors to the adapter training.}
  \label{fig:prel2_app}
\end{figure}

\begin{figure}[hbt!]
  \centering
  \includegraphics[width=.85\linewidth]{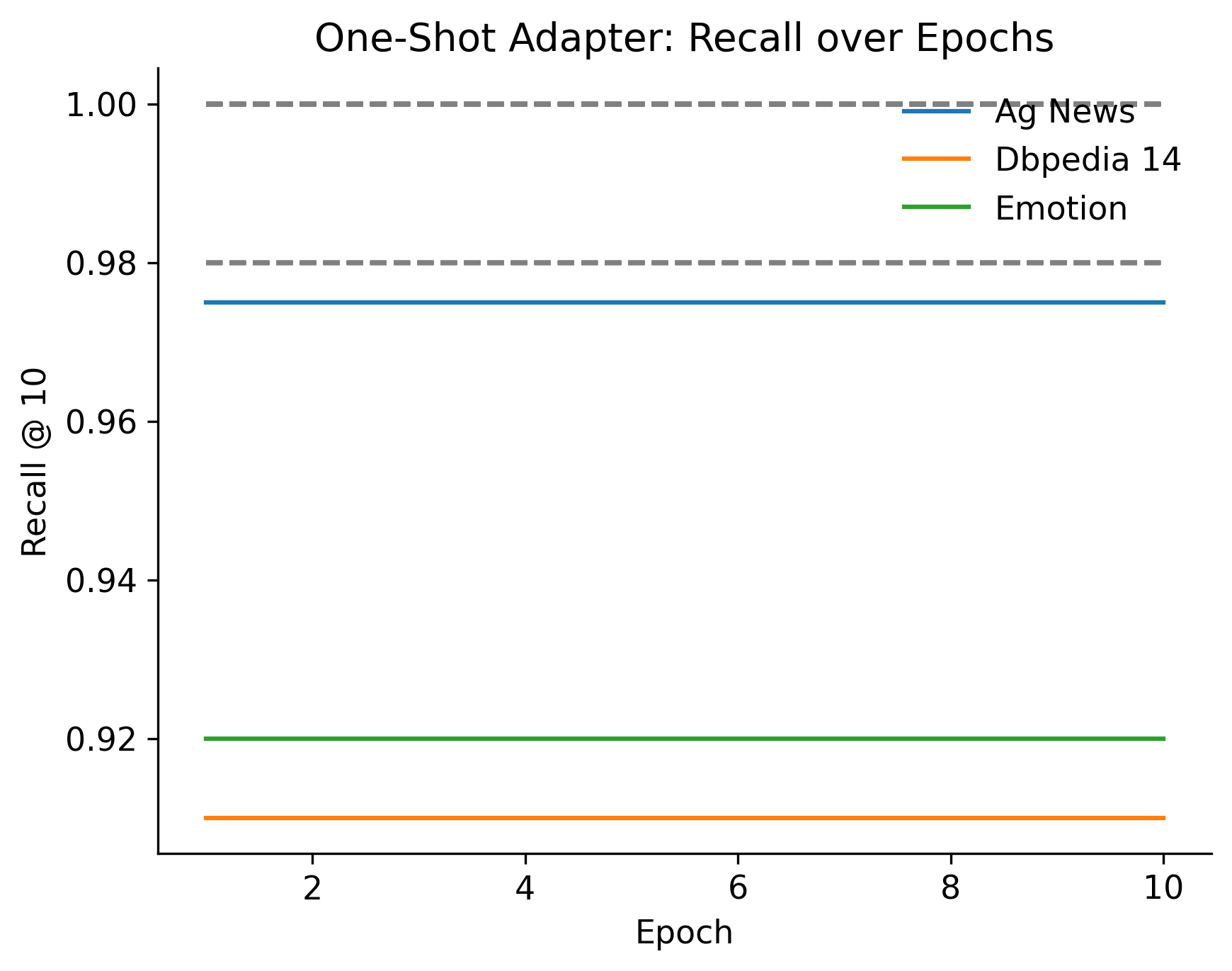}
  \caption{One-shot (closed-form SVD solution) OP fitting vs. multi-epoch SGD optimisation for the Orthogonal Procrustes loss on AG-News R@10 ARR (from initial explorations). Iterative optimization (e.g., 2-5 epochs with SGD) can sometimes yield slightly better results than the direct one-shot SVD solution, though the difference is usually small.}
  \label{fig:one-shot_app}
\end{figure}

\end{document}